\documentclass{article}

\usepackage{arxiv}

\usepackage[utf8]{inputenc} 
\usepackage[T1]{fontenc}    
\usepackage{hyperref}       
\usepackage{url}            
\usepackage{booktabs}       
\usepackage{amsfonts}       
\usepackage{nicefrac}       
\usepackage{microtype}      
\usepackage{lipsum}
\usepackage{natbib}
\usepackage{multirow}
\usepackage{graphicx}
\usepackage{amsmath}
\usepackage{framed} 
\usepackage[figuresright]{rotating}
\usepackage{times}
\usepackage{latexsym}
\usepackage{makecell}
\usepackage{multirow}
\usepackage{graphicx}
\usepackage{url}
\usepackage{color} 
\usepackage{natbib}
\definecolor{shadecolor}{rgb}{0.92,0.92,0.92} 
\usepackage{lineno}
\graphicspath{ {./images/} }

\title{Bridging Coarse and Fine Recognition: A Hybrid Approach for Open-Ended Multi-Granularity Object Recognition in Interactive Educational Games}

\author{
 Hanling Yi \\
  Intellifusion Inc.\\
   \And
 Feng Lin \\
  Intellifusion Inc.\\
  \And
 Mao Luo \\
  Intellifusion Inc.\\
  \And
 Yifan Yang \\
  Intellifusion Inc. \\
  \And
 Xiaotian Yu \\
  Intellifusion Inc. \\
  \And
 Rong Xiao \\
  Intellifusion Inc.
}

\begin{document}
\maketitle
\begin{abstract}
Recent advances in Multimodal Large Language Models (MLLMs) have enabled open-ended object recognition, yet they struggle with fine-grained tasks. In contrast, CLIP-style models excel at fine-grained recognition but lack broad coverage of general object categories. To bridge this gap, we propose \textbf{HyMOR}, a \textbf{Hy}brid \textbf{M}ulti-granularity open-ended \textbf{O}bject \textbf{R}ecognition framework that integrates an MLLM with a CLIP model. In HyMOR, the MLLM performs open-ended and coarse-grained object recognition, while the CLIP model specializes in fine-grained identification of domain-specific objects such as animals and plants. This hybrid design enables accurate object understanding across multiple semantic granularities, serving as a robust perceptual foundation for downstream multi-modal content generation and interactive gameplay. To support evaluation in content-rich and educational scenarios, we introduce TBO (TextBook Objects), a dataset containing 20,942 images annotated with 8,816 object categories extracted from textbooks. Extensive experiments demonstrate that HyMOR narrows the fine-grained recognition gap with CLIP to 0.2\% while improving general object recognition by 2.5\% over a baseline MLLM, measured by average Sentence-BERT (SBert) similarity. Overall, HyMOR achieves a 23.2\% improvement in average SBert across all evaluated datasets, highlighting its effectiveness in enabling accurate perception for multi-modal game content generation and interactive learning applications. 
\end{abstract}


\section{Introduction}
Recent advances in Multimodal Large Language Models (MLLMs) have shown impressive capability in open-ended object recognition~\cite{yang2024qwen2,liu2023visual,li2023blip}, opening new opportunities in the field of AI for educational applications. These models can function as virtual tutors, assisting children in recognizing and understanding objects in their everyday environments. Motivated by this potential, we develop a lightweight AI perception system designed for AI-powered educational cameras, where children can capture photos of objects in their surroundings and receive AI-generated, interactive feedback. 

In such an AI camera product, recognized objects serve as the foundation for downstream multi-modal content generation, including storytelling, bilingual (e.g., Chinese–English) language learning, pronunciation guidance, and functional explanations, all driven by the recognized objects. To further encourage engagement, the AI camera incorporates a game-like reward mechanism, in which users earn virtual coins through interactions and use them to nurture a virtual pet, forming a closed-loop learning and gameplay experience. By combining visual perception with interactive feedback and game mechanics, the AI camera supports preschool children in developing early language skills and cognitive understanding in an intuitive and immersive manner.

In open-world educational contexts, object recognition task requires two key properties: open-endedness and multi-level semantic granularity. On one hand, it is impractical to predefine a closed vocabulary covering all objects encountered in daily life, especially in unconstrained environments such as homes, classrooms, parks, or zoos. This necessitates open-ended recognition models capable of generating object labels beyond a fixed category set~\citep{LI2024110258}. On the other hand, different application scenarios demand varying levels of recognition granularity~\cite{YU2026111955}. For example, coarse-grained recognition is sufficient for many everyday objects (e.g., LEGO Brick, Backpack), whereas fine-grained recognition of animals or plants (e.g., Bellis Perennis, Reticulated Giraffe) can provide significantly richer educational value for children. Based on these considerations, we categorize objects into two distinct groups:

1) General objects: Common objects for which coarse-grained recognition suffices. Due to their long-tail distribution, exhaustive labeling is impractical, posing challenges in open-world scenarios.

2) Specialized objects: Educationally valuable objects that require fine-grained recognition. While they are easier to enumerate~\citep{stevens2024bioclip}, distinguishing subtle inter-class differences remains a key challenge. 

\begin{figure}[t]
\centering
\includegraphics[width=3in]{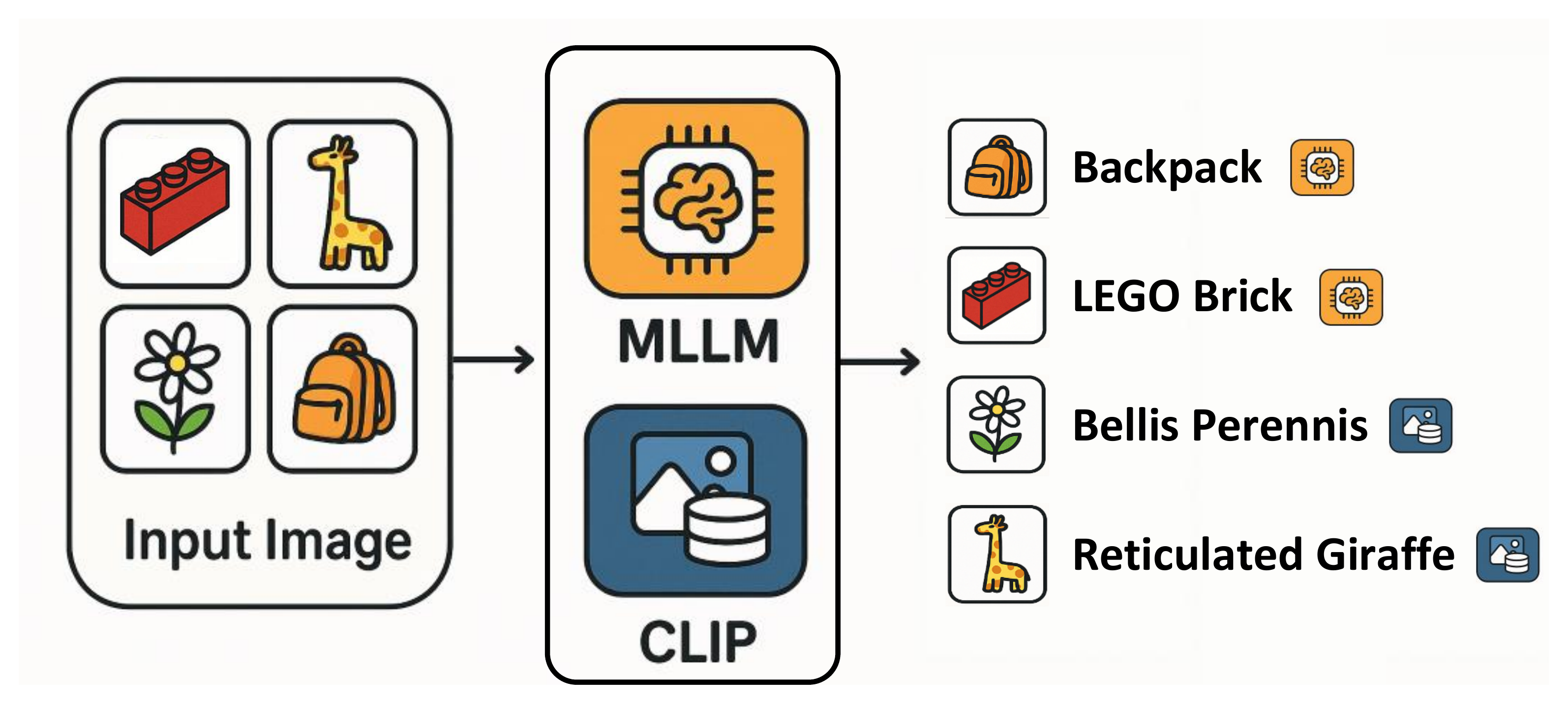}
\caption{Overview of the proposed HyMOR framework, which integrates an MLLM and a CLIP model. The MLLM functions as both a router and a coarse-grained recognizer for general objects, while the CLIP model provides fine-grained recognition for specialized objects, such as animals and plants.}
\label{fig:overview}
\end{figure}

The primary challenge lies in designing a multi-granularity, open-ended object recognition model that can effectively balance accuracy between coarse-grained recognition of general objects and fine-grained recognition of specialized objects. In this study, we define specialized objects as plant and animal species\footnote{We focus on plant and animal species because children frequently encounter them in zoos or parks, where recognizing these species offers meaningful educational benefits.}. While current MLLMs perform well in coarse-grained recognition, their fine-grained performance remains limited as compared to CLIP models~\citep{zhangvisually,geigle2024african,heanalyzing}. However, CLIP models are not inherently open-ended and typically lack the coverage required for open-world scenarios. These limitations make neither approach alone well-suited for interactive educational systems deployed on AI cameras.

\begin{figure}[t]
\centering
\includegraphics[width=3.0in]{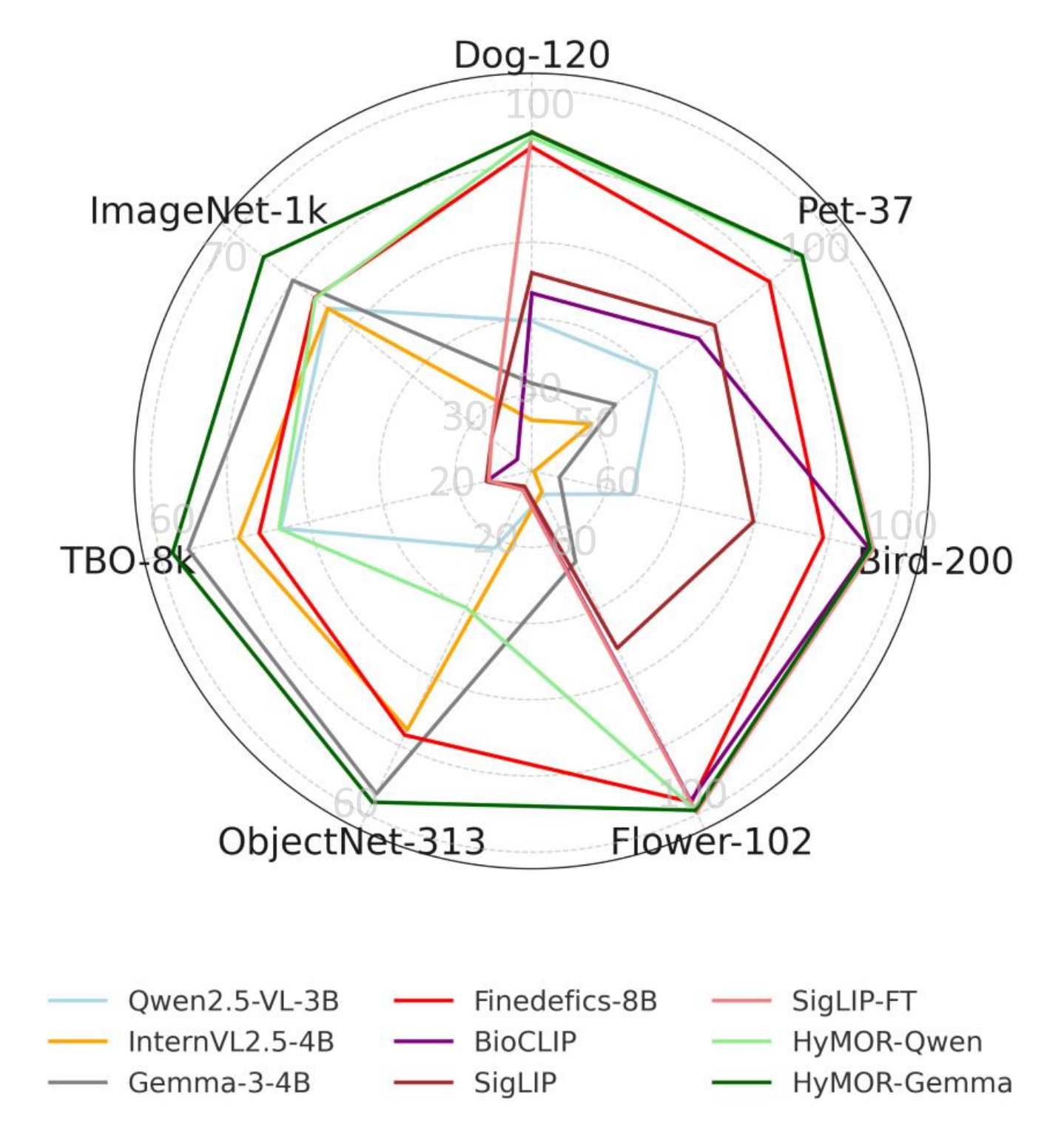}
\caption{Sentence-BERT similarity scores of different models across fine-grained datasets (Dog-120, Pet-37, Bird-200, and Flower-102) and coarse-grained datasets (ImageNet-1K, ObjectNet-313, and TBO-8K). Our proposed HyMOR framework, represented by HyMOR-Gemma, achieves the best overall performance among all evaluated models.}
\label{fig:radar}
\end{figure}

To overcome these limitations, we propose HyMOR, a simple yet effective hybrid framework that integrates an MLLM~\citep{liu2023visual} with a CLIP model~\citep{radford2021learning}, as shown in Figure~\ref{fig:overview}. The MLLM serves as a router and a coarse-grained general object recognizer, while the CLIP handles fine-grained recognition for specialized objects. Specifically, given an input image, the MLLM generates both the general object category (e.g., animal, plant, or other) and the coarse object name (e.g., Backpack, LEGO Brick, or Flower, etc). If the object belongs to the specialized objects category (e.g., animal or plant), the CLIP is activated to extract the image feature, which is then used to retrieve the most relevant label from a pre-processed vector database containing embeddings of various species of animals and plants. If the similarity score of the retrieved result exceeds a pre-defined threshold, the system adopts the fine-grained label from the retrieval result. Otherwise, it defaults to the coarse label generated by the MLLM. This design enables HyMOR to maintain open-ended coverage while selectively applying fine-grained recognition when it is most beneficial.

To complement existing object recognition benchmarks, we introduce an education oriented test set named TBO (TextBook Objects). TBO is designed to cover fundamental visual concepts commonly found in textbooks, which consists of 20,942 images with 8,816 object names extracted from textbooks.  
Extensive experiments across both fine-grained and coarse-grained recognition tasks demonstrates the effectiveness of HyMOR (Figure~\ref{fig:radar}). Specifically, HyMOR-Gemma achieves a 23.2\% improvement in average SBert across all datasets over Gemma-3-4B. It narrows the fine-grained performance gap to SigLIP-FT (our fine-tuned variant of SigLIP~\citep{zhai2023sigmoid}, which achieves the best fine-grained performance among CLIP-based models in our experiments) to just 0.2\%, while improving general recognition by 2.5\% over the MLLM baseline, both measured by average SBert score. 

To summary, the contributions of this paper are as follows:
\begin{itemize}
    \item We propose HyMOR, a hybrid framework that combines the open-ended world knowledge of an MLLM with the fine-grained visual representations of a CLIP model, enabling multi-granularity object recognition for open-world educational application.
    \item We construct an education-oriented benchmark named TBO, consisting of 20,942 images with 8,816 object names extracted from school textbooks, to fill the gap in benchmarks for educational AI.  
    \item We conduct comprehensive experiments showing that HyMOR improves average SBert by 23.2\% across all datasets over baseline. Specifically, it narrows the fine-grained recognition gap to CLIP model to just 0.2\%, improves general recognition by 2.5\% over the baseline MLLM.
\end{itemize}

\section{Related Work}

\textbf{Fine-Grained Visual Recognition (FGVR).}  
FGVR is a longstanding challenge in computer vision that focuses on distinguishing subordinate-level categories within visually similar object classes~\citep{liu2024democratizing, tang2022learning, QI201947}. Due to the subtle inter-class differences, FGVR methods typically rely on powerful feature representation models such as CLIP to enhance discrimination of objects~\citep{ZHAO2022108618,SHAN2022108748}. For instance, BioCLIP~\citep{stevens2024bioclip} combines CLIP-style multimodal contrastive learning~\citep{radford2021learning} with hierarchical biological taxonomies to improve fine-grained classification in biological domains. Other approaches aim to strengthen the FGVR capabilities of MLLMs by incorporating open-set classification data during fine-tuning~\citep{geigle2024african}, or by employing contrastive learning strategies over object-attribute-category triples~\citep{heanalyzing}.

\noindent\textbf{MLLMs for Image Classification.}  
MLLMs align pre-trained image encoders, typically a Vision Transformer (ViT) from CLIP, with a Large Language Model (LLM), enabling them to understand both images and text. Recent research has revisited the image classification task within the MLLM paradigm. For example, Zhang et. al.~\cite{zhangvisually} observed that MLLMs substantially underperform CLIP on standard image classification benchmarks such as ImageNet. In contrast, Liu et. al.~\cite{liu2024revisiting} demonstrated that some modern MLLMs can match or even surpass CLIP-style models, attributing these gains to advancements in LLM architectures and the diversification of training data sources. Nevertheless, a substantial performance gap remains between MLLMs and CLIP models in fine-grained tasks, especially in recognizing animal and plant species.

Our work builds upon these insights by introducing HyMOR, a hybrid framework that combines the coarse-grained recognition strength of MLLMs with the fine-grained retrieval capabilities of CLIP models, effectively bridging the performance gap across recognition granularities.

\section{Problem and Preliminary Result}
\subsection{Multi-granularity Open Ended Object Recognition}

In this paper, we address the task of open-ended object recognition, where the goal is to generate the name of the main object depicted in a given input image. The main object is defined as the most visually salient entity, typically distinguished by attributes such as size, position, color, or shape that set it apart from the background.

Open-ended object recognition differs fundamentally from traditional image classification. While both tasks aim to identify the salient object in an image, conventional image classification operates under a closed-set assumption, where the set of possible categories is fixed and known during training. In contrast, open-ended object recognition is an open-world task that requires the model to recognize and describe objects without relying on predefined categories—even for objects it has never encountered during training. This setting resembles open-world object detection~\citep{joseph2021towards, wang2023detecting} but is simpler—since our task does not require localizing the object. 

Furthermore, our target application imposes an additional requirement: the model must support multi-granularity recognition. Specifically, for specialized objects such as animals and plants, we aim for fine-grained recognition at the species level, which provides greater educational value. For general objects (e.g., furniture or household items), coarse-grained labels are sufficient. This  requirement introduces new challenges in model design, demanding both generalization and precision across varying semantic granularities.

\subsection{Preliminary Result}
\label{sec:preliminary_result}

\begin{table*}[tp]
\centering
\begin{tabular}{r | c | c | c | c | c | c | c | c}
\hline
\multirow{2}{*}{\textbf{Model}} & \multicolumn{2}{c |}{\textbf{Dog-120}} & \multicolumn{2}{c |}{\textbf{Bird-200}} & \multicolumn{2}{c |}{\textbf{Flower-102}} & \multicolumn{2}{c}{\textbf{ImageNet-1k}}\\
\cline{2-9}
& EM & SBert  & EM  & SBert & EM &   SBert & EM &  SBert \\
\hline
Baseline  & $26.7$ & $63.6$  & $21.5$ & $63.6$ & $20.3$ & $53.4$ & $18.7$ & $55.8$ \\
Finetuned &  $73.8 \uparrow$ & $85.6\uparrow$  & $5.2\downarrow$ & $48.0\downarrow$ & $12.8\downarrow$ & $46.2\downarrow$ & $14.1\downarrow$ & $49.4\downarrow$\\
\hline
\end{tabular}
\caption{Preliminary results on three FGVR datasets and ImageNet-1K. ``Baseline'' denotes the Qwen2.5-VL-3B-Instruct model, while ``Finetuned'' refers to its fine-tuned version which is trained exclusively on Dog-120 and evaluated across all datasets. We report both Exact Match (EM) and Sentence-BERT similarity (SBert) scores. Arrows indicate performance changes: “$\uparrow$” denotes an improvement on the in-domain test set, while “$\downarrow$” indicates a decrease on out-of-distribution (OOD) test sets.}
\label{tab:result_preliminary}
\end{table*}

A straightforward way to enhance the fine-grained recognition capability of existing MLLMs is to fine-tune them on FGVR datasets. However, our preliminary experiments reveal a critical trade-off: while fine-tuning boosts performance on in-domain test sets, it significantly degrades performance on out-of-domain datasets, particularly for object classes unseen during training.

To illustrate this, we fine-tune Qwen2.5-VL-3B-Instruct~\citep{Qwen2.5-VL} on the Stanford dog-120~\citep{Krause_Stark_Deng_Fei-Fei_2013} FGVR dataset, which includes 120 dog breeds and 12,000 training samples. Following prior work~\citep{geigle2024african}, we augment the training set with 24,000 general Visual Question Answering (VQA) samples randomly sampled from LLaVA-1.5-665k~\citep{liu2024improved}, aiming to preserve the model’s general QA capabilities and mitigate overfitting. The model is fine-tuned for 2 epochs using a learning rate of 2e-5. We then evaluate the fine-tuned model on two additional FGVR datasets—Bird-200 and Flower-102—as well as the coarse-grained ImageNet-1K dataset. 

As shown in Table~\ref{tab:result_preliminary}, fine-tuning leads to a substantial performance gain on the in-domain Dog-120 test set. However, it causes a marked decline in performance on other fine-grained datasets such as Bird-200 and Flower-102, as well as on the coarse-grained ImageNet-1K dataset. These findings indicate that naively fine-tuning MLLMs on a single fine-grained dataset can compromise their general object recognition capabilities. Motivated by this observation, our proposed method avoids fine-tuning the MLLM on any FGVR dataset. Instead, we freeze the MLLM and restrict fine-tuning to the CLIP model, thereby preserving the MLLM’s broad generalization ability while enhancing fine-grained recognition through retrieval.

\section{Methodology}

\begin{figure*}[!ht]
\centering
\includegraphics[width=4.8in]{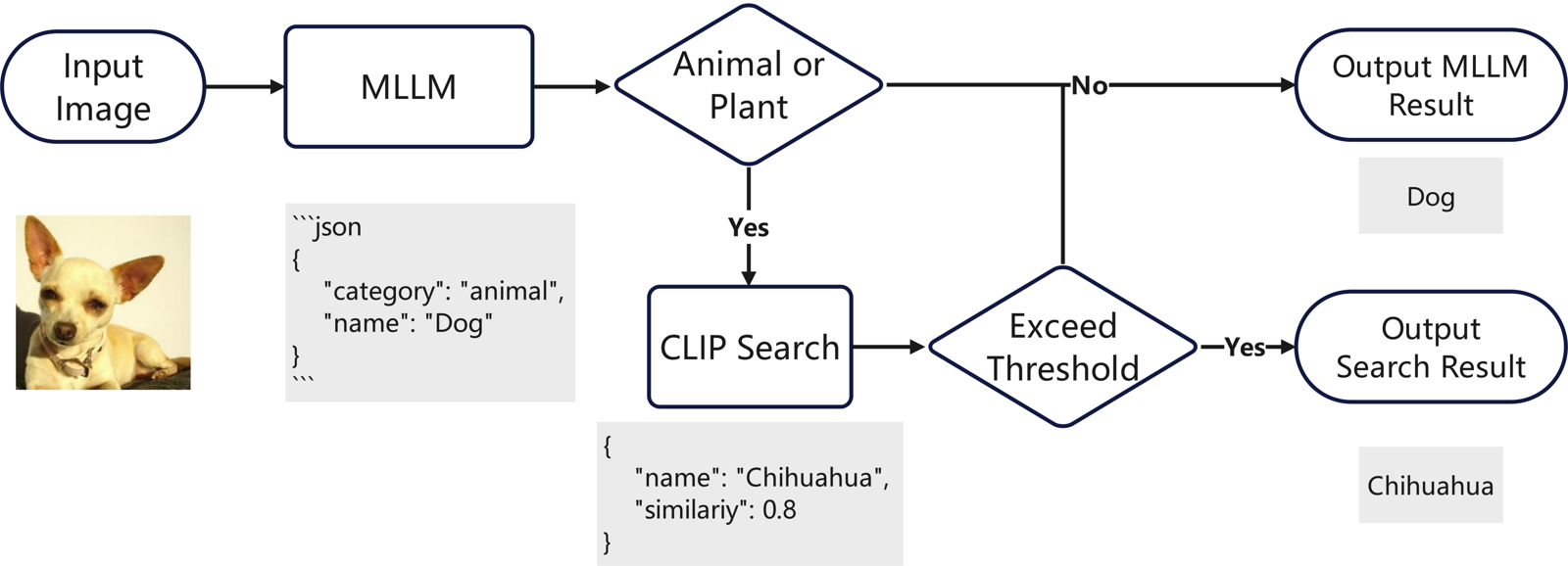}
\caption{An illustrative example of our proposed HyMOR framework for multi-granularity open-ended object recognition. Given an input image, the MLLM predicts the object’s general category (e.g., animal, plant, or other) and a coarse label (e.g., dog). For specialized categories (animal or plant), the CLIP-ViT extracts image features to retrieve the closest match from a pre-processed vector database. If the search similarity exceeds a predefined threshold, the fine-grained label is returned; otherwise, the system defaults to the MLLM's coarse label.}
\label{fig:overall}
\end{figure*}

To achieve balanced performance across both coarse-grained and fine-grained recognition tasks while maintaining open-endedness, we propose \textbf{HyMOR} (Hybrid Multi-granularity open-ended Object Recognition), a framework that integrates an MLLM with a CLIP model. Figure~\ref{fig:overall} illustrates the overall framework of HyMOR. Below, we detail its key components and their respective roles.

\subsection{MLLM as Router and Coarse-Grained Recognizer} 
In HyMOR, the MLLM plays a dual role, functioning both as a router and as a coarse-grained general object recognition engine. For each input image, the MLLM generates: (1) a general category classification (animal, plant, or other), and (2) a coarse object name. We implement this using open-source MLLMs with carefully designed prompts, as detailed in~\ref{sec:append_prompt}. 

If the MLLM classifies the object as ``other'', the system directly outputs the coarse object name predicted by the MLLM. However, if the object is categorized as  ``animal'' or ``plant'', the fine-grained recognition pipeline is subsequently activated. It is important to note that we do not fine-tune the MLLM on any FGVR dataset, thus preserving its generalization capabilities.

\subsection{CLIP-Based Fine-Grained Recognition}
For fine-grained recognition, HyMOR leverages the vision encoder from the CLIP model. In our implementation, we fine-tune the CLIP model on both the FGVR training set and a randomly sampled subset of TreeOfLife-10M, as will be described in Section~\ref{sec:exp}. Following the approach of BioCLIP~\citep{stevens2024bioclip}, we precompute class centroids by averaging the image embeddings for each class in the training data and store these centroids in a vector database for efficient retrieval. 

During inference, the ViT in CLIP extracts image embeddings from the input. We then compute similarity scores between the query embedding and all stored class centroids. If the similarity score of the nearest centroid exceeds a predefined confidence threshold, the system outputs the corresponding fine-grained label. Otherwise, it defaults to the MLLM’s coarse-grained prediction.

This hierarchical design ensures computational efficiency and robustness across varying recognition granularities. By selectively engaging the fine-grained pipeline only when necessary and falling back to the MLLM’s coarse-grained prediction when confidence is low, HyMOR maintains high accuracy while remaining lightweight and suitable for edge deployment.

\section{Textbooks Dataset}
To complement existing object recognition benchmarks, we construct a test dataset named TBO (\textbf{T}ext\textbf{B}ook \textbf{O}bjects), comprising object names extracted from educational textbooks. This dataset is specifically designed to assess recognition of educationally relevant objects that students are expected to learn, making it particularly relevant to assess models for real-world educational applications.

\begin{figure}[ht]
\centering
\includegraphics[width=2.8in]{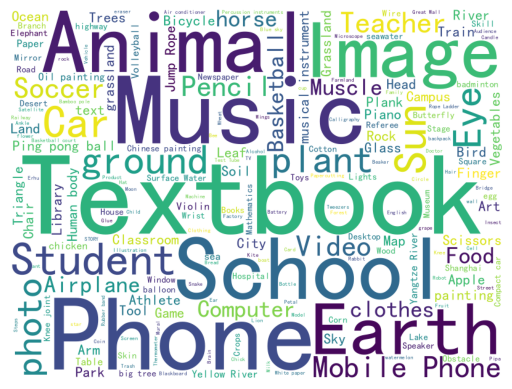}
\caption{Word cloud of object names in the TBO dataset, sized according to their frequency in textbooks.}
\label{fig:wordcloud}
\end{figure}

Our dataset construction process involves three key steps: First, we obtain textbooks from a widely recognized national educational platform\footnote{Smart Education of China: \url{https://basic.smartedu.cn/tchMaterial}}, which hosts PDF versions of textbooks across all subjects from primary to high school. In total, we gather 2,623 PDF files (1,284 from primary school, 834 from secondary school, and 505 from high school). We then employ PyMuPDF4LLM\footnote{\url{https://pymupdf.readthedocs.io/en/latest/pymupdf4llm/}} for PDF parsing and use Qwen2.5-72B-Instruct~\citep{qwen2.5} to extract commonly occurring object names and, when necessary, translate them from other languages into English. Finally, we gather representative images for each object using a text-to-image search engine\footnote{\url{https://image.baidu.com/}}, followed by human annotations. On average, each object name has 2.4 images in TBO. Detailed statistics of the TBO dataset are summarized in Table~\ref{tab:statics_tbo}. We visualize the object names based on their frequency in the textbook in Figure~\ref{fig:wordcloud}.

\begin{table}[h]
\centering
\begin{tabular}{r | c | c | c | c}
\hline
& Primary & Secondary & High & Total \\
\hline
PDFs & 1284 & 834 & 505 & 2623 \\
Objects & 5392 & 2011 & 1413 & 8816 \\
Images & 12778 & 4812 & 3352 & 20942 \\
\hline
\end{tabular}
\caption{The statistics of TBO test dataset.}
\label{tab:statics_tbo}
\end{table}

\section{Experiments}

\subsection{Training and Evaluation Details}
\label{sec:exp}
We conduct experiments on four widely used fine-grained visual recognition (FGVR) datasets: Caltech-UCSD Bird-200~\citep{Caltech_UCSD}, Stanford Dog-120~\citep{Krause_Stark_Deng_Fei-Fei_2013}, Oxford-IIIT Pet-37~\citep{Parkhi_Vedaldi_Zisserman_Jawahar_2012}, and Flower-102~\citep{Nilsback_Zisserman_2008}. The training sets from these FGVR datasets are used to train the CLIP component of our HyMOR framework. 

Additionally, prior research~\citep{pham2023combined} has shown that scaling both dataset size and batch size can enhance the generalization ability of CLIP models. Following this insight, we augment our training data with a curated subset of TreeOfLife-10M~\citep{stevens2024bioclip}, the largest machine learning-ready biological image dataset to date, containing over 10 million images spanning 454,000 taxonomic labels. For our experiments, we randomly sample 300,000 images representing 149,071 distinct species to enrich our training set and enhance the model's fine-grained recognition of animals and plants.

Our CLIP implementation builds upon SigLIP~\citep{zhai2023sigmoid}, specifically adopting SigLIP-SO400M-Patch14-384 (denoted as SigLIP) as the vision encoder, paired with a 64-token causal autoregressive transformer as the text encoder. We train the model for 35 epochs using a cosine learning rate schedule~\citep{loshchilov2016sgdr} and denote the resulting model as SigLIP-FT. Training is distributed across 32 NVIDIA A800-80GB GPUs, with a global batch size of 512. During inference, we compute class centroids by averaging image embeddings from all training samples (excluding the TreeOfLife-10M subset). After applying mean subtraction and L2 normalization to both centroids and test feature vectors, predictions are made by selecting the class whose centroid is nearest to the test embedding. 

For the MLLM component in HyMOR, we utilize two open-source models: Qwen2.5-VL-3B~\citep{Qwen2.5-VL} and Gemma-3-4B~\citep{gemma_2025}, referred to as HyMOR-Qwen and HyMOR-Gemma respectively in our experiments (see Table~\ref{tab:result_fgvr}). Notably, we freeze the MLLMs in HyMOR and employ carefully designed prompt to guide MLLMs to generate both the category and the name of the main object in the image, as detailed in~\ref{sec:append_prompt}. For comparison, we also benchmark our method against general-purpose MLLMs such as InternVL2.5-4B~\citep{chen2024expanding}, as well as fine-grained MLLMs like Finedefics-8B~\citep{heanalyzing}.

Beyond the FGVR test sets, we evaluate model performance on several coarse-grained datasets, including ImageNet-1K, ObjectNet-313, as well as our curated education-oriented dataset, TBO-8K. The statistics of test sets are summarized in Table~\ref{tab:statics_testset}.

\begin{table}[h]
\centering
\begin{tabular}{r | c | c | c }
\hline
Dataset & Gran. & Classes & Samples  \\
\hline
Dog-120 &\multirow{4}{*}{Fine} & 120 & 8580 \\
Pet-37 & & 37 & 3669 \\
Birds-200 & & 200 & 5794 \\
Flower-102 & & 102 & 6149 \\ 
\hline
ImageNet-1K & \multirow{3}{*}{Coarse} & 1000 & 50000  \\
ObjectNet-313 & & 313 & 50273 \\
TBO-8K & & 8816 & 20942 \\
\hline
\end{tabular}
\caption{Statistics of the evaluation datasets. ``Gran.'' is short for ``Granularity''.}
\label{tab:statics_testset}
\end{table}

For evaluation metrics, we primarily report exact match (EM) accuracy and the semantic similarity between the ground truth and predicted labels, computed using Sentence-BERT\footnote{We use all-mpnet-base-v2 from Sentence Transformers.} (SBert). Additionally, given the inherent ambiguity of open-ended object recognition, we employ GPT-4 as an external judger to assess semantic alignment between predictions and ground truth. Further details of the LLM-as-a-judge setup are provided in ~\ref{app:llm_judge}.

\subsection{Results}

\begin{table}[!htp]
\centering
\scalebox{0.83}{
\begin{tabular}{r | c | c | c | c | c | c | c | c | c | c}
\hline
\multirow{2}{*}{\textbf{Model}} & \multicolumn{2}{c |}{\textbf{Dog-120}} & \multicolumn{2}{c |}{\textbf{Pet-37}} & \multicolumn{2}{c |}{\textbf{Bird-200}} & \multicolumn{2}{c |}{\textbf{Flower-102}} & \multirow{2}{*}{\makecell[c]{Avg. \\ EM}} & \multirow{2}{*}{\makecell[c]{Avg. \\ SBert}} \\
\cline{2-9}
& EM & SBert  &  EM & SBert & EM  & SBert & EM &   SBert & \\
\hline
\multicolumn{9}{c}{\textit{General MLLMs}} \\
\hline
Qwen2.5-VL-3B  & 26.7 & 63.6 & 34.3 &  65.1 & 21.5 & 63.6 & 20.3 & 53.4 & 25.7 & 61.4 \\
InternVL2.5-4B &  5.0 & 48.0 & 16.4 & 51.9 & 4.8 & 50.3 & 16.2 & 53.0 & 10.6 & 50.8\\
Gemma-3-4B & 11.8 & 53.8 &  27.0 & 56.8 & 2.1 & 53.7 & 27.0 & 63.2 & 17.0 & 56.9\\
\hline
\multicolumn{9}{c}{\textit{Fine-grained MLLMs}} \\
\hline
Finedefics-8B & 79.8 & 91.0 & 66.9 & 87.8 & 74.5 & 89.2 & 92.6 & 98.2 & 78.5 & 91.6\\
\hline
\multicolumn{9}{c}{\textit{CLIP-based}} \\
\hline
BioCLIP & 44.4 & 68.0 & 54.4 & 73.5 & 88.7 & 95.3 & 94.1 & 98.0 & 70.4 & 83.7\\
SigLIP & 50.7 & 71.2 & 57.8 & 76.8 & 51.9 & 79.8 & 63.6& 75.8 & 56.0 & 75.9 \\
SigLIP-FT & \textbf{85.8} & \textbf{93.4} & \textbf{84.4} & \textbf{94.4} & \textbf{89.7} & \textbf{95.9} & \textbf{96.5} & \textbf{99.7} & \textbf{89.1} & \textbf{95.9}\\
\hline
\multicolumn{9}{c}{\textit{HyMOR framework}} \\
\hline
HyMOR-Qwen & 84.7 & 92.5 & 84.1 & 94.2 & 89.0 & 95.6 & 95.9 & 99.3 & 88.4 & 95.4\\
HyMOR-Gemma & 85.7 & 93.3 & \textbf{84.4} & \textbf{94.4} & 88.8 & 95.5 & 96.0 & 99.4 & 88.7 & 95.7\\
\hline
\end{tabular}
}
\caption{The experimental results on four FGVR datasets. We show the exact match score (EM) and Sentence BERT similarity (SBert) for each datasets.}
\label{tab:result_fgvr}
\end{table}

\begin{table*}[htp] 
\centering
\scalebox{0.83}{
\begin{tabular}{r | c | c | c | c | c | c | c | c | c | c }
\hline
\multirow{2}{*}{\textbf{Model}} & \multicolumn{3}{c |}{\textbf{ImageNet-1k}} & \multicolumn{3}{c|}{\textbf{ObjectNet-313}} & \multicolumn{3}{c|}{\textbf{TBO-8k}} & \multirow{2}{*}{\makecell[c]{Avg. \\ SBert}} \\
\cline{2-10}
& EM & SBert  & LLM & EM & SBert & LLM & EM  & SBert & LLM \\
\hline
\multicolumn{10}{c}{\textit{General MLLMs}} \\
\hline
Qwen2.5-VL-3B  & 18.7 & 55.8 & 45.7 & 1.8 & 25.1 &  4.1 & 11.1
& 45.4 & 35.9 & 42.1\\
InternVL2.5-4B &  15.5 & 55.8 & 54.7 & 14.7 & 48.9 & 31.7 & 8.7 & 50.5 & 38.2  & 51.7\\
Gemma-3-4B & 19.3 & 61.1 & 55.5 & 22.7 & 57.3 & 45.6 & 16.2 & 56.6 & \textbf{55.6} & 58.3 \\
\hline
\multicolumn{10}{c}{\textit{Fine-grained MLLMs}} \\
\hline
Finedefics-8B & 22.3 & 57.8 & 31.8 & 16.8 & 49.6 & 25.7 & 13.2 & 48.0 & 21.1 & 51.8 \\
\hline
\multicolumn{10}{c}{\textit{CLIP-based}} \\
\hline
BioCLIP & 0.5 & 27.2 & 7.3 & 0.0 & 17.4 & 0.1 & 0.4 & 20.5 & 1.0 & 21.7\\
SigLIP & 0.5 & 31.1 & 8.6 & 0.0 & 17.0 & 0.1 & 0.4 & 20.5 & 1.0 & 22.9\\
SigLIP-FT & 0.7 & 31.1 & 11.8 & 0.0 & 17.5 & 0.1  & 0.6 & 20.2 & 1.3 & 22.9\\
\hline
\multicolumn{10}{c}{\textit{HyMOR Framework}} \\
\hline
HyMOR-Qwen & 21.9 & 57.7 & 49.5 & 5.1 & 32.9 & 12.0 & 10.4 & 45.6 & 36.5 & 45.4\\
HyMOR-Gemma & \textbf{26.9} & \textbf{65.5} & \textbf{60.9} & \textbf{24.4} & \textbf{58.4} & \textbf{46.9} & \textbf{19.8} & \textbf{58.5} & 54.6 &\textbf{60.8}\\
\hline
\end{tabular}
}
\caption{The experimental results on three general object recognition datasets. We show the exact match score (EM), Sentence BERT similarity (SBert) and LLM-Judge score (LLM) for each datasets.}
\label{tab:result_general}
\end{table*}

\begin{figure*}[ht]
\centering
\includegraphics[width=4.8in]{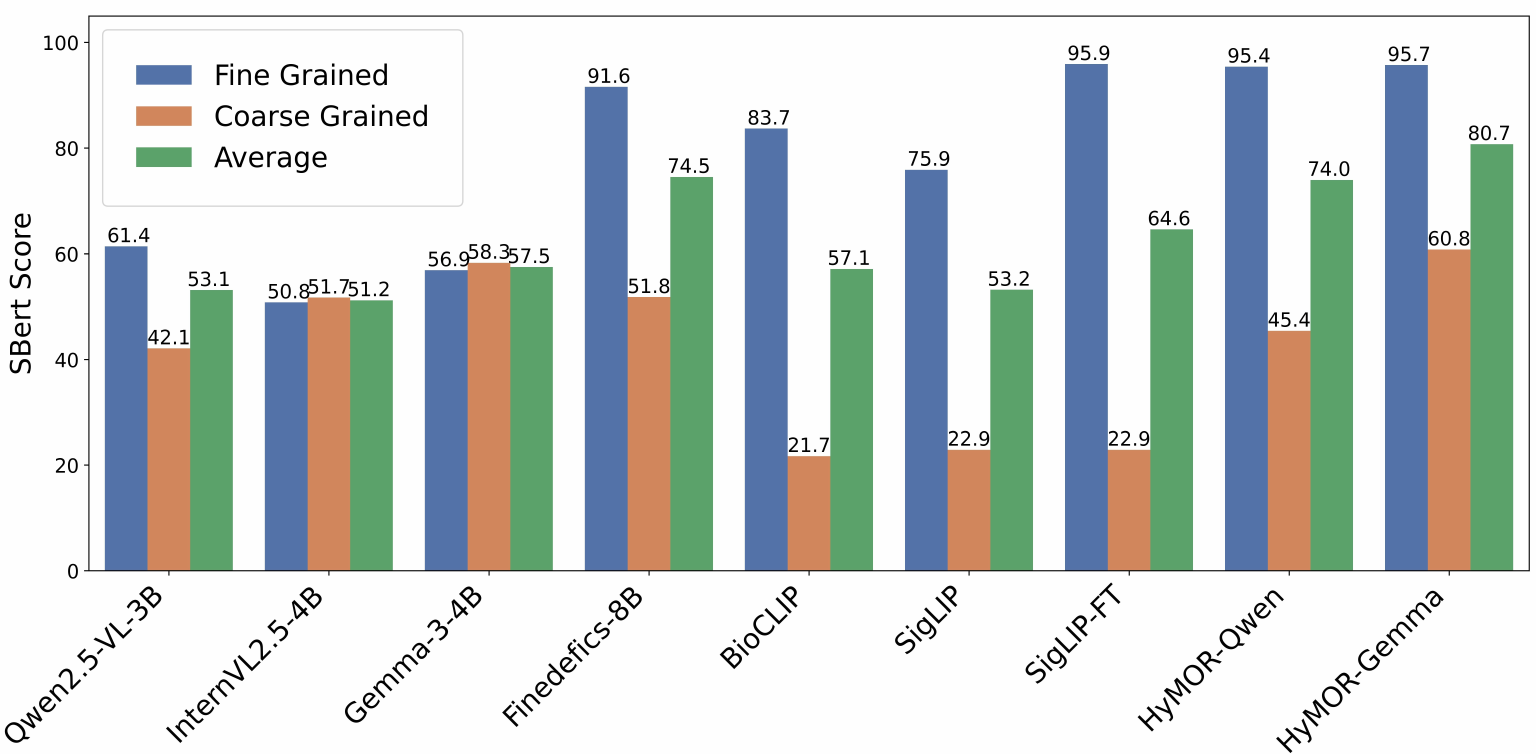}
\caption{The average Sentence-Bert scores of different
models across fine-grained, coarse-grained and all datasets. }
\label{fig:sbert}
\end{figure*}


\noindent\textbf{Fine-Grained Tasks.} As shown in Table~\ref{tab:result_fgvr}, CLIP-based models deliver the best performance on fine-grained recognition. In particular, SigLIP-FT, which is specifically fine-tuned on FGVR datasets, achieves the highest accuracy across all four benchmarks, with an average EM score of 89.1 and consistently high SBert similarity scores exceeding 93.4. In contrast, general-purpose MLLMs perform poorly on these tasks, reaffirming prior findings that such models struggle with fine-grained recognition~\citep{zhangvisually}. For example, Qwen2.5-VL-3B attains only an average EM of 25.7, while InternVL2.5-4B falls to 10.6. The fine-grained MLLM, Finedefics-8B, shows notable improvement over general MLLMs, achieving an average EM score of 78.5, but still falls short of SigLIP-FT.

Our proposed HyMOR framework substantially enhances fine-grained recognition performance. Both HyMOR-Qwen and HyMOR-Gemma exhibit significant improvements over their respective MLLM baselines. Specifically, HyMOR-Qwen achieves an average SBert of 95.4, while HyMOR-Gemma reaches 95.7, effectively closing much of the gap to SigLIP-FT. We attribute the small remaining performance difference to potential misclassifications during the initial MLLM categorization stage: since HyMOR activates the CLIP-based fine-grained module only when the MLLM categorizes the object as an animal or plant, any errors at this stage may limit recall and overall recognition accuracy.

\noindent\textbf{Coarse-Grained Tasks.} Moving to the coarse-grained benchmarks (Table~\ref{tab:result_general}), HyMOR-Gemma consistently achieves the best overall performance. For instance, it records 26.9 EM, 65.5 SBert, and 60.9 LLM-Judge scores on ImageNet-1K, and 24.4 EM, 58.4 SBert, and 46.9 LLM-Judge on ObjectNet-313. Notably, MLLMs outperform CLIP-based models by a substantial margin on coarse-grained tasks. For example, SigLIP-FT achieves only 0.7 EM on ImageNet-1K and 0.6 EM on TBO-8K, whereas Qwen2.5-VL-3B achieves 18.7 EM and 11.1 EM, respectively. This disparity is expected, as the CLIP vector database primarily covers fine-grained FGVR categories and lacks representation of the broader object spectrum required for general object recognition. By contrast, MLLMs benefit from extensive pretraining on diverse datasets, enabling stronger generalization.

Our HyMOR framework delivers modest yet meaningful gains over the MLLM baselines in these coarse-grained scenarios. For instance, HyMOR-Qwen improves upon Qwen2.5-VL-3B by +3.3 average SBert, while HyMOR-Gemma surpasses Gemma-3-4B by 2.5 in average SBert. These improvements primarily arise from animal and plant categories within the coarse-grained datasets, where the CLIP component of HyMOR continues to contribute effectively.

\noindent\textbf{MLLM Routing Accuracy.} We evaluate the routing accuracy of the MLLM component in HyMOR-Gemma. We note that all objects in the FGVR datasets belong to specialized categories (animals or plants), while all objects in ObjectNet-313 fall under the general category. Accordingly, we measure the routing accuracy for specialized objects across the four FGVR datasets and for general objects on ObjectNet-313. HyMOR-Gemma achieves a routing accuracy of 99.8\% for specialized objects and 95.6\% for general objects. These results demonstrate the high reliability of the MLLM in effectively directing inputs to the appropriate recognition branch.

\noindent\textbf{Summary.} We show the average SBert scores of different models across fine-grained, coarse-grained and all datasets in Figure~\ref{fig:sbert}. Overall, these results demonstrate that HyMOR effectively balances accuracy across both fine-grained and coarse-grained recognition tasks. In particular, HyMOR-Gemma achieves a 23.2\% improvement in average SBert score across all datasets over Gemma-3-4B (80.7\% VS 57.5\%). It narrows the performance gap to 0.2\% with SigLIP-FT on FGVR datasets while maintaining competitive performance on general object recognition benchmarks (+2.5\% improvement over Gemma-3-4B), validating its effectiveness as a unified, lightweight solution for multi-granularity open-ended object recognition.

\section{Conclusion}



In this study, we address the challenge of multi-granularity, open-ended object recognition in interactive educational settings that require both coarse-grained coverage and fine-grained semantic understanding. We introduce HyMOR, a hybrid framework that integrates a MLLM for open-ended, coarse-grained object recognition with a CLIP-based model for fine-grained identification of specialized objects. Extensive experimental results validate the effectiveness of HyMOR, highlighting its suitability as a robust perceptual foundation for real-world educational systems.

Beyond standalone recognition performance, HyMOR enables multi-modal content generation and interactive gameplay by providing accurate and semantically rich object understanding. In particular, the framework supports downstream applications such as AI-powered educational cameras, where recognized objects drive narrative generation, language learning, and game-like interactions. As such, HyMOR contributes to the broader goal of facilitating intelligent, multi-modal, and engaging content generation for educational and game-based experiences.

Looking forward, incorporating grounding and localization capabilities to identify object regions within images would further enhance the system’s interactivity and usefulness. Additionally, extending the framework from single salient object recognition to multi-object and scene-level understanding represents a promising direction for future work, enabling richer world modeling and more complex multi-modal content generation in interactive game and learning environments.

\appendix
\section{Details of LLM-as-a-Judge}
\label{app:llm_judge}
We employ a large language model as an automated judge to assess the consistency between predicted object names and ground truth annotations. For each input image with a ground truth label (GT), the evaluated model produces a predicted label (PRED). A prompt is then constructed using GT and PRED based on a predefined template. Upon receiving the prompt, the judge outputs one of three possible ratings: ``A'', ``B'', or ``C'', corresponding to scores of 1.0, 0.8, and 0.0, respectively. The average score across all test images serves as the performance metric for the dataset. We use OpenAI’s GPT API as the judge, specifically the ``gpt-4-0613'' model with API version ``2024-08-01-preview''. For reference, the prompt template is presented below.

\vspace{0.1cm} 
\begin{minipage}[b][1\totalheight][c]{0.92\columnwidth}%
\begin{shaded}%
Given two English nouns, prediction and label, determine their semantic relationship. Output A if they are synonyms (i.e., they refer to the same concept). Output B if either prediction is a hypernym of label, or label is a hypernym of prediction. Output C if neither of the above applies. Respond with a single uppercase letter: A, B, or C. 

Examples:

prediction = car, label = automobile → Output: A; 

prediction = vegetable, label = bokchoy → Output: B; 

prediction = bokchoy, label = vegetable → Output: B; 

prediction = airplane, label = football → Output: C. 

Now evaluate: prediction = \{PRED\}, label = \{GT\}
\end{shaded}%
\end{minipage}
\vspace{0.1cm} 

\section{Prompts for MLLM}
\label{sec:append_prompt}
We use the following prompt to guide open-source MLLM to generate both the category and the name of the main object in the image:

\vspace{0.1cm} 
\begin{minipage}[b][1\totalheight][c]{0.92\columnwidth}%
\begin{shaded}%
Identify the main object in the image. Output the category and the name of the main object in the image in JSON format with the keys "category" and "name". The category can only be one of "plant", "animal" or "other".

Example output:

\{"category": "animal", "name": "Dog"\}

Now generate the output for the given image.
\end{shaded}%
\end{minipage}
\vspace{0.1cm} 

Additionally, to enhance output stability, we apply guided decoding in vLLM~\citep{kwon2023efficient}.

\bibliographystyle{unsrt}  
\bibliography{custom}  

@article{liu2023visual,
  title={Visual instruction tuning},
  author={Liu, Haotian and Li, Chunyuan and Wu, Qingyang and Lee, Yong Jae},
  journal={Advances in neural information processing systems},
  volume={36},
  pages={34892--34916},
  year={2023}
}

@article{yang2024qwen2,
  title={Qwen2. 5 technical report},
  author={Yang, An and Yang, Baosong and Zhang, Beichen and Hui, Binyuan and Zheng, Bo and Yu, Bowen and Li, Chengyuan and Liu, Dayiheng and Huang, Fei and Wei, Haoran and others},
  journal={arXiv preprint arXiv:2412.15115},
  year={2024}
}

@inproceedings{li2023blip,
  title={Blip-2: Bootstrapping language-image pre-training with frozen image encoders and large language models},
  author={Li, Junnan and Li, Dongxu and Savarese, Silvio and Hoi, Steven},
  booktitle={International conference on machine learning},
  pages={19730--19742},
  year={2023},
  organization={PMLR}
}

@inproceedings{liu2024democratizing,
  title={DEMOCRATIZING FINE-GRAINED VISUAL RECOGNITION WITH LARGE LANGUAGE MODELS},
  author={Liu, M and Roy, S and Li, W and Zhong, Z and Sebe, N and Ricci, E and others},
  booktitle={12th International Conference on Learning Representations, ICLR 2024},
  year={2024},
  organization={International Conference on Learning Representations, ICLR}
}

@inproceedings{stevens2024bioclip,
  title={Bioclip: A vision foundation model for the tree of life},
  author={Stevens, Samuel and Wu, Jiaman and Thompson, Matthew J and Campolongo, Elizabeth G and Song, Chan Hee and Carlyn, David Edward and Dong, Li and Dahdul, Wasila M and Stewart, Charles and Berger-Wolf, Tanya and others},
  booktitle={Proceedings of the IEEE/CVF conference on computer vision and pattern recognition},
  pages={19412--19424},
  year={2024}
}

@inproceedings{radford2021learning,
  title={Learning transferable visual models from natural language supervision},
  author={Radford, Alec and Kim, Jong Wook and Hallacy, Chris and Ramesh, Aditya and Goh, Gabriel and Agarwal, Sandhini and Sastry, Girish and Askell, Amanda and Mishkin, Pamela and Clark, Jack and others},
  booktitle={International conference on machine learning},
  pages={8748--8763},
  year={2021},
  organization={PmLR}
}

@inproceedings{zhangvisually,
  title={Why are Visually-Grounded Language Models Bad at Image Classification?},
  author={Zhang, Yuhui and Unell, Alyssa and Wang, Xiaohan and Ghosh, Dhruba and Su, Yuchang and Schmidt, Ludwig and Yeung-Levy, Serena},
  booktitle={The Thirty-eighth Annual Conference on Neural Information Processing Systems},
  year=2024
}

@inproceedings{geigle2024african,
  title={African or European Swallow? Benchmarking Large Vision-Language Models for Fine-Grained Object Classification},
  author={Geigle, Gregor and Timofte, Radu and Glava{\v{s}}, Goran},
  booktitle={Proceedings of the 2024 Conference on Empirical Methods in Natural Language Processing},
  pages={2653--2669},
  year={2024}
}

@inproceedings{heanalyzing,
  title={Analyzing and Boosting the Power of Fine-Grained Visual Recognition for Multi-modal Large Language Models},
  author={He, Hulingxiao and Li, Geng and Geng, Zijun and Xu, Jinglin and Peng, Yuxin},
  booktitle={The Thirteenth International Conference on Learning Representations},
  year={2025}
}

@article{Caltech_UCSD,
  title={The Caltech-UCSD Birds-200-2011 Dataset},
  author={ Wah, Catherine  and  Branson, Steve  and  Welinder, Peter  and  Perona, Pietro  and  Belongie, Serge },
  journal={california institute of technology},
  year={2011},
}

@inproceedings{Krause_Stark_Deng_Fei-Fei_2013,  
 title={3D Object Representations for Fine-Grained Categorization}, 
 url={http://dx.doi.org/10.1109/iccvw.2013.77}, 
 DOI={10.1109/iccvw.2013.77}, 
 booktitle={2013 IEEE International Conference on Computer Vision Workshops}, 
 author={Krause, Jonathan and Stark, Michael and Deng, Jia and Fei-Fei, Li}, 
 year={2013}, 
 month={Dec}, 
 language={en-US} 
 }

@inproceedings{Parkhi_Vedaldi_Zisserman_Jawahar_2012,  
 title={Cats and dogs}, 
 url={http://dx.doi.org/10.1109/cvpr.2012.6248092}, 
 DOI={10.1109/cvpr.2012.6248092}, 
 booktitle={2012 IEEE Conference on Computer Vision and Pattern Recognition}, 
 author={Parkhi, O. M. and Vedaldi, A. and Zisserman, A. and Jawahar, C. V.}, 
 year={2012}, 
 month={Jun}, 
 language={en-US} 
 }

@inproceedings{Nilsback_Zisserman_2008,  
 title={Automated Flower Classification over a Large Number of Classes}, 
 url={http://dx.doi.org/10.1109/icvgip.2008.47}, 
 DOI={10.1109/icvgip.2008.47}, 
 booktitle={2008 Sixth Indian Conference on Computer Vision, Graphics \& Image Processing}, 
 author={Nilsback, Maria-Elena and Zisserman, Andrew}, 
 year={2008}, 
 month={Dec}, 
 language={en-US} 
 }

@article{liu2024revisiting,
  title={Revisiting mllms: An in-depth analysis of image classification abilities},
  author={Liu, Huan and Xiao, Lingyu and Liu, Jiangjiang and Li, Xiaofan and Feng, Ze and Yang, Sen and Wang, Jingdong},
  journal={arXiv preprint arXiv:2412.16418},
  year={2024}
}

@article{pham2023combined,
  title={Combined scaling for zero-shot transfer learning},
  author={Pham, Hieu and Dai, Zihang and Ghiasi, Golnaz and Kawaguchi, Kenji and Liu, Hanxiao and Yu, Adams Wei and Yu, Jiahui and Chen, Yi-Ting and Luong, Minh-Thang and Wu, Yonghui and others},
  journal={Neurocomputing},
  volume={555},
  pages={126658},
  year={2023},
  publisher={Elsevier}
}

@inproceedings{zhai2023sigmoid,
  title={Sigmoid loss for language image pre-training},
  author={Zhai, Xiaohua and Mustafa, Basil and Kolesnikov, Alexander and Beyer, Lucas},
  booktitle={Proceedings of the IEEE/CVF international conference on computer vision},
  pages={11975--11986},
  year={2023}
}

@article{loshchilov2016sgdr,
  title={Sgdr: Stochastic gradient descent with warm restarts},
  author={Loshchilov, Ilya and Hutter, Frank},
  journal={arXiv preprint arXiv:1608.03983},
  year={2016}
}

@article{qwen2.5,
    title   = {Qwen2.5 Technical Report}, 
    author  = {An Yang and Baosong Yang and Beichen Zhang and Binyuan Hui and Bo Zheng and Bowen Yu and Chengyuan Li and Dayiheng Liu and Fei Huang and Haoran Wei and Huan Lin and Jian Yang and Jianhong Tu and Jianwei Zhang and Jianxin Yang and Jiaxi Yang and Jingren Zhou and Junyang Lin and Kai Dang and Keming Lu and Keqin Bao and Kexin Yang and Le Yu and Mei Li and Mingfeng Xue and Pei Zhang and Qin Zhu and Rui Men and Runji Lin and Tianhao Li and Tingyu Xia and Xingzhang Ren and Xuancheng Ren and Yang Fan and Yang Su and Yichang Zhang and Yu Wan and Yuqiong Liu and Zeyu Cui and Zhenru Zhang and Zihan Qiu},
    journal = {arXiv preprint arXiv:2412.15115},
    year    = {2024}
}

@article{Qwen2.5-VL,
  title={Qwen2.5-VL Technical Report},
  author={Bai, Shuai and Chen, Keqin and Liu, Xuejing and Wang, Jialin and Ge, Wenbin and Song, Sibo and Dang, Kai and Wang, Peng and Wang, Shijie and Tang, Jun and Zhong, Humen and Zhu, Yuanzhi and Yang, Mingkun and Li, Zhaohai and Wan, Jianqiang and Wang, Pengfei and Ding, Wei and Fu, Zheren and Xu, Yiheng and Ye, Jiabo and Zhang, Xi and Xie, Tianbao and Cheng, Zesen and Zhang, Hang and Yang, Zhibo and Xu, Haiyang and Lin, Junyang},
  journal={arXiv preprint arXiv:2502.13923},
  year={2025}
}

@inproceedings{liu2024improved,
  title={Improved baselines with visual instruction tuning},
  author={Liu, Haotian and Li, Chunyuan and Li, Yuheng and Lee, Yong Jae},
  booktitle={Proceedings of the IEEE/CVF Conference on Computer Vision and Pattern Recognition},
  pages={26296--26306},
  year={2024}
}

@article{gemma_2025,
    title={Gemma 3},
    url={https://goo.gle/Gemma3Report},
    publisher={Kaggle},
    author={Gemma Team},
    year={2025}
}

@article{chen2024expanding,
  title={Expanding Performance Boundaries of Open-Source Multimodal Models with Model, Data, and Test-Time Scaling},
  author={Chen, Zhe and Wang, Weiyun and Cao, Yue and Liu, Yangzhou and Gao, Zhangwei and Cui, Erfei and Zhu, Jinguo and Ye, Shenglong and Tian, Hao and Liu, Zhaoyang and others},
  journal={arXiv preprint arXiv:2412.05271},
  year={2024}
}

@inproceedings{kwon2023efficient,
  title={Efficient Memory Management for Large Language Model Serving with PagedAttention},
  author={Woosuk Kwon and Zhuohan Li and Siyuan Zhuang and Ying Sheng and Lianmin Zheng and Cody Hao Yu and Joseph E. Gonzalez and Hao Zhang and Ion Stoica},
  booktitle={Proceedings of the ACM SIGOPS 29th Symposium on Operating Systems Principles},
  year={2023}
}

@article{tang2022learning,
  title={Learning attention-guided pyramidal features for few-shot fine-grained recognition},
  author={Tang, Hao and Yuan, Chengcheng and Li, Zechao and Tang, Jinhui},
  journal={Pattern Recognition},
  volume={130},
  pages={108792},
  year={2022},
  publisher={Elsevier}
}

@article{QI201947,
title = {Exploiting spatial relation for fine-grained image classification},
journal = {Pattern Recognition},
volume = {91},
pages = {47-55},
year = {2019},
issn = {0031-3203},
doi = {https://doi.org/10.1016/j.patcog.2019.02.007},
url = {https://www.sciencedirect.com/science/article/pii/S0031320319300718},
author = {Lei Qi and Xiaoqiang Lu and Xuelong Li},
}

@inproceedings{joseph2021towards,
  title={Towards open world object detection},
  author={Joseph, KJ and Khan, Salman and Khan, Fahad Shahbaz and Balasubramanian, Vineeth N},
  booktitle={Proceedings of the IEEE/CVF conference on computer vision and pattern recognition},
  pages={5830--5840},
  year={2021}
}

@inproceedings{wang2023detecting,
  title={Detecting everything in the open world: Towards universal object detection},
  author={Wang, Zhenyu and Li, Yali and Chen, Xi and Lim, Ser-Nam and Torralba, Antonio and Zhao, Hengshuang and Wang, Shengjin},
  booktitle={Proceedings of the IEEE/CVF Conference on Computer Vision and Pattern Recognition},
  pages={11433--11443},
  year={2023}
}

@article{ZHAO2022108618,
title = {A feature consistency driven attention erasing network for fine-grained image retrieval},
journal = {Pattern Recognition},
volume = {128},
pages = {108618},
year = {2022},
issn = {0031-3203},
doi = {https://doi.org/10.1016/j.patcog.2022.108618},
url = {https://www.sciencedirect.com/science/article/pii/S0031320322000991},
author = {Qi Zhao and Xu Wang and Shuchang Lyu and Binghao Liu and Yifan Yang}
}

@article{SHAN2022108748,
title = {Self-Attention based fine-grained cross-media hybrid network},
journal = {Pattern Recognition},
volume = {130},
pages = {108748},
year = {2022},
issn = {0031-3203},
doi = {https://doi.org/10.1016/j.patcog.2022.108748},
url = {https://www.sciencedirect.com/science/article/pii/S0031320322002291},
author = {Wei Shan and Dan Huang and Jiangtao Wang and Feng Zou and Suwen Li},
}

@article{LI2024110258,
title = {Learning adversarial semantic embeddings for zero-shot recognition in open worlds},
journal = {Pattern Recognition},
volume = {149},
pages = {110258},
year = {2024},
issn = {0031-3203},
doi = {https://doi.org/10.1016/j.patcog.2024.110258},
url = {https://www.sciencedirect.com/science/article/pii/S0031320324000098},
author = {Tianqi Li and Guansong Pang and Xiao Bai and Jin Zheng and Lei Zhou and Xin Ning},
}

@article{YU2026111955,
title = {Structural feature enhanced transformer for fine-grained image recognition},
journal = {Pattern Recognition},
volume = {169},
pages = {111955},
year = {2026},
issn = {0031-3203},
doi = {https://doi.org/10.1016/j.patcog.2025.111955},
url = {https://www.sciencedirect.com/science/article/pii/S0031320325006156},
author = {Ying Yu and Wei Wei and Cairong Zhao and Jin Qian and Enhong Chen},
}



\end{document}